\pdfoutput=1
\documentclass[11pt]{article}
\newcommand{\jocelyn}[1]{ {\color{red!50!black}$_\text{js}$} }
\newcommand{\yubin}[1]{ {\color{blue!50!black}$_\text{yk}$} }
\newcommand{\sharifa}[1]{ {\color{green!50!black}$_\text{sa}$} }
\newcommand{\mohit}[1]{ {\color{orange!50!black}$_\text{mh}$} }

\newcommand{\kt}[1]{ {\color{red!50!black}$_\text{KT}$} }

\usepackage{amssymb}
\usepackage{times}
\usepackage{latexsym}
\usepackage{graphics}
\usepackage{tabularx}
\usepackage{amsmath}
\usepackage{pifont}
\usepackage{tabularx}
\usepackage{multirow} 
\usepackage{arydshln}
\usepackage[dvipsnames]{xcolor}
 \usepackage{subcaption}
\usepackage{lipsum,graphicx}
\usepackage{cuted}
\usepackage{capt-of,etoolbox}
\usepackage{hyperref}
\usepackage{cleveref}

\usepackage[]{ACL2023}

\newcommand{\cmark}{\text{\ding{51}}}
\newcommand{\xmark}{\text{\ding{55}}}
\usepackage[T1]{fontenc}
\usepackage[utf8]{inputenc}

\usepackage{microtype}

\usepackage{inconsolata}

\title{EmpathicStories++: A Multimodal Dataset for Empathy towards Personal Experiences}

\author{
    Jocelyn Shen$^{*}$\enspace\quad
    Yubin Kim$^{*}$ \quad
    \textbf{Mohit Hulse} \quad
    \textbf{Wazeer Zulfikar}\quad \\ \textbf{Sharifa Alghowinem} \quad \textbf{Cynthia Breazeal} \quad \textbf{Hae Won Park}\\
    \normalsize{Massachusetts Institute of Technology, Cambridge, MA, USA} \\
    \normalsize{\texttt{\{joceshen, ybkim95, mhulse, wazeer, sharifah,  cynthiab, haewon\}@mit.edu}}
}

\begin{document}
\makeatletter
\apptocmd\@maketitle{{\myfigure{}\par}}{}{}
\makeatother
% \begin{document}
\newcommand\myfigure{%
\vspace{-20pt}
\centering
\includegraphics[width=\textwidth]{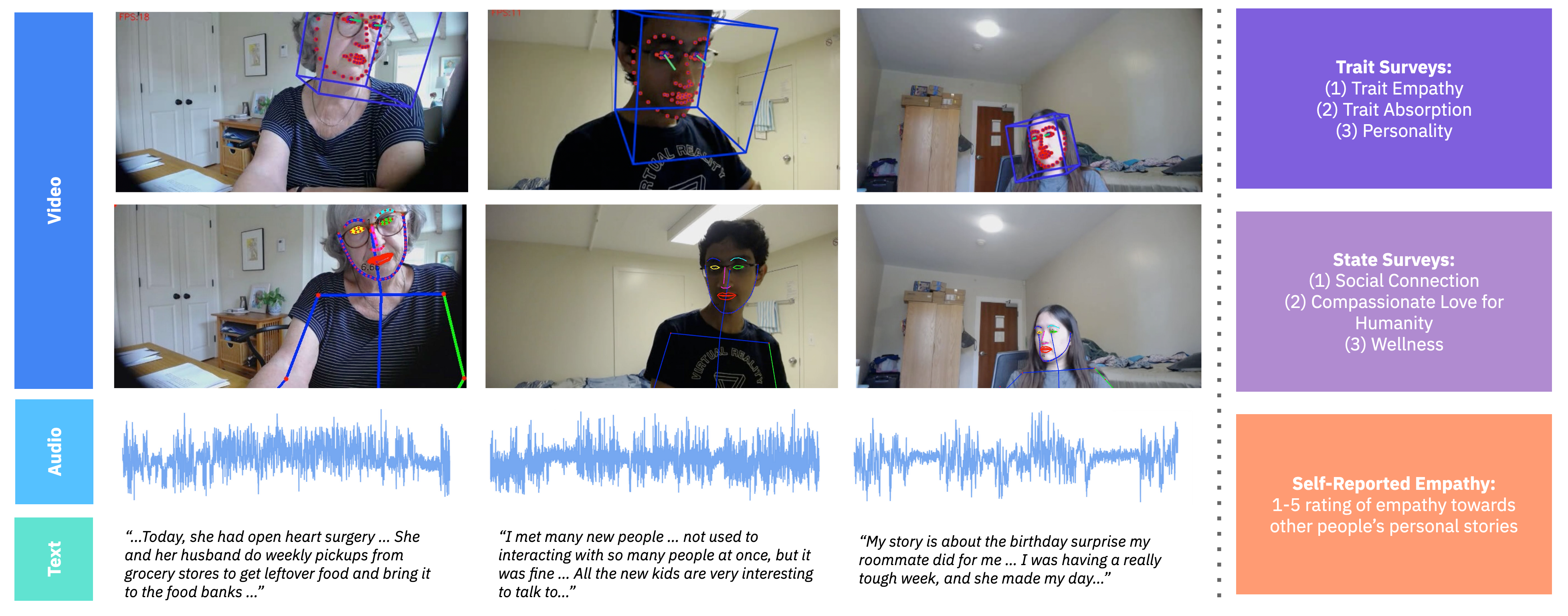}
% Hae Won Park: minor comment on text boxes - pay attention to accessible color palettes when choosing text + background color combinations
\captionof{figure}{\textbf{The \textsc{EmpathicStories++} dataset }is collected from a month-long in-the-wild deployment of 41 participants (across 269 sessions with 53 hours of data) telling personal stories and reading other people's personal stories with an AI agent. We publicly release all video, audio, and text data in addition to psychometric surveys in order to advance computational empathy research, and more broadly, social-emotional reasoning in AI.}
\vspace{5pt}

\label{fig:teaser}
}
\maketitle

% \begin{figure*}[ht]
%     \centering
%     \includegraphics[width=\textwidth]{figures/datasetoverview.png}
%     \caption{\textbf{The \textsc{EmpathicStories++} dataset }is collected from a month-long in-the-wild deployment of 41 participants (across 269 sessions with 53 hours of data) telling personal stories and reading other people's personal stories with an AI agent. We publicly release all video, audio, and text data in addition to psychometric surveys in order to advance computational empathy research, and more broadly, social-emotional reasoning in AI.}

%     \label{fig:teaser}
% \end{figure*}

\begin{abstract}
\vspace{-5pt}
Modeling empathy is a complex endeavor that is rooted in interpersonal and experiential dimensions of human interaction, and remains an open problem within AI. Existing empathy datasets fall short in capturing the richness of empathy responses, often being confined to in-lab or acted scenarios, lacking longitudinal data, and missing self-reported labels. We introduce a new multimodal dataset for empathy during personal experience sharing: the \textsc{EmpathicStories++} dataset\footnote{\url{https://mitmedialab.github.io/empathic-stories-multimodal/}} containing 53 hours of video, audio, and text data of 41 participants sharing vulnerable experiences and reading empathically resonant stories with an AI agent. \textsc{EmpathicStories++} is the first longitudinal dataset on empathy, collected over a month-long deployment of social robots in participants' homes, as participants engage in natural, empathic storytelling interactions with AI agents. 
We then introduce a novel task of predicting individuals' empathy toward others' stories based on their personal experiences, evaluated in two contexts: participants' own personal shared story context and their reflections on stories they read. We benchmark this task using state-of-the-art models to pave the way for future improvements in contextualized and longitudinal empathy modeling. Our work provides a valuable resource for further research in developing empathetic AI systems and understanding the intricacies of human empathy within genuine, real-world settings.
% \kt{Looks great now, focuses on ++s, shares why it is valuable to the AI community and beyond.}
% In summary, "EmpathicStories++" sets a new benchmark as the first multimodal, in-the-wild, longitudinal dataset for understanding empathy, enabling the development of empathetic AI systems and fostering research into the intricacies of human empathy within genuine, real-world settings.
\end{abstract}

\section{Introduction}
% Kush's comments: one thing to think about: 
% - who are the type of reviewers that will read this? 
% - what would they look for/knit pick? can we solve those cases directly, tailor the story? 
\label{sec:intro}

Empathy is a fundamental pillar of interpersonal human  interactions ranging from prosocial behavior to enhancing human connection \cite{morelli_emerging_2015}.
Modeling and understanding empathy is a complex task, due to its inherently interpersonal and experiential nature: empathy is tied to neurological synchronizations between representations of self and other \cite{decety_human_nodate}, and is dependent on a person's past experiences \cite{hodges_giving_2010}. 
Interest in empathy within AI communities has grown in recent years, as systems advance in context-awareness, naturalness, and fluency, although they typically fall short in social reasoning \cite{sap_neural_2022}. 
Few prior works present datasets that are sufficient to capture the richness of human empathy responses during personal experience sharing. These datasets are limited in the following ways: (1) They are not captured in-the-wild. Existing multimodal empathy datasets are sourced from in-lab, online, or acted settings, which may differ greatly from empathy expressed in natural conversations. (2) They are not longitudinal, capturing only one-shot interaction settings, despite empathy being dependent on a combination of many past experiences. (3) Previous datasets are not self-labeled. While empathy can be inferred by external cues, it is an inherently subjective process, requiring self-reported labels for user-centric or personalized modeling.

\begin{table*}[ht]
    \centering
\def\arraystretch{1.25}%  1 is the default, change whatever you need
    % \footnotesize
    \resizebox{\textwidth}{!}{%
\begin{tabular}{c|c|c|c|c|c|c|c|c}
\hline 

\textbf{Dataset} 
& \textbf{Modalities} 
& \textbf{Self-annotated} 
& \textbf{Longitudinal} 
& \textbf{Source} 
& \textbf{Collected in-the-wild}  
& \textbf{\# Subjects}
& \textbf{Quantity (video/audio)}
& \textbf{Quantity (text)}
% & \multicolumn{2}{c|}{\textbf{Quantity of data}} 
\\ 
\hline\hline

MELD \cite{poria_meld_2019}
& $\mathrm{V} + \mathrm{A} + \mathrm{T}$
& $\xmark$ 
& $\xmark$ 
& TV 
& $\xmark$ 
% & 
% 6 emotion categories (Joy, 
% Sadness, Fear, Anger, Surprise, 
% and Disgust)
& 407 
& $-$
& 13,708 utterances 
\\ 
% \hline 

M3ED \cite{zhao_m3ed_2022}
& $\mathrm{V} + \mathrm{A} + \mathrm{T}$
& $\xmark$ 
& $\xmark$ 
& TV 
& $\xmark$ 
% & 
% 7 emotion categories (happy, 
% surprise, sad, disgust, anger, 
% fear, neutral) 
& 626 
& $-$ 
& 990 dialogues $/$ 24,449 utterances
\\ 
% \hline 

Emolnt-MD \cite{singh_emoint-trans_2023}
& $\mathrm{V} + \mathrm{A} + \mathrm{T}$
& $\xmark$ 
& $\xmark$ 
& Movies 
& $\xmark$
% & 
% 32 emotions and 15 empathetic 
% intents
& 4375 
& 534 hrs $/$ 32,040 min
& 724,756 utterances
\\ 
% \hline 

MEDIC \cite{zhouan_zhu_medic_2023}
& $\mathrm{V} + \mathrm{A} + \mathrm{T}$
& $\xmark$ 
& $\xmark$ 
& Acted motivational interviews 
& $\xmark$ 
% & 
% expression of experience, 
% emotional reactions, cognitive 
% reactions 
& $-$ 
& 11 hrs $/$ 678 min 
& 771 utterances
\\ 
% \hline 

OMG-Empathy \cite{barros_omg-empathy_2019}
& $\mathrm{V} + \mathrm{A} + \mathrm{T}$ 
& $\cmark$
& $\xmark$ 
& In-lab 
& $\xmark$ 
% & valence 
& 10 listeners, 2 speakers 
& 8 hrs $/$ 480 min 
& $-$ 
\\ 
% \hline 

EmpatheticDialogues \cite{rashkin_towards_2019}
& $\mathrm{T}$ 
& $\xmark$ 
& $\xmark$ 
& Crowdsourced 
& $\xmark$ 
% & 
% 32 emotions and 8 empathetic 
% response intents 
& 810 
& $-$ 
& 24,850 dialogues $/$ 107,247 utterances 
\\ 
% \hline 

Empathic Conversations \cite{omitaomu_empathic_2022}
& $\mathrm{T}$ 
& $\cmark$ 
& $\xmark$ 
& Crowdsourced 
& $\xmark$
% & 
% personality and demographic 
% info, perceived empathy, 
% turn-level annotations of 
% empathy, self-disclosure, 
% emotion 
& 92 
& $-$ 
& 5,821 utterances
\\ 
% \hline 

EmpathicStories \cite{shen_modeling_2023}
& $\mathrm{T}$ 
& $\xmark$ 
& $\xmark$ 
& Online stories 
& $\cmark$ 
% & 
% summaries of event, emotion, 
% and moral, and empathic 
% similarity between 2 stories 
& $-$ 
& $-$ 
& 1,500 stories 

\\ 
% \hline 

\citet{sharma_computational_2020}
& $\mathrm{T}$ 
& $\xmark$ 
& $\xmark$ 
& Online peer support platforms 
& $\cmark$ 

% & 
% emotional reactions, 
% interpretations, and explorations 
& $-$ 
& $-$ 
& 10,143 utterances
\\ 
% \hline 

EDOS \cite{welivita_large-scale_2021}
& $\mathrm{T}$ 
& $\xmark$ 
& $\xmark$ 
& Movie subtitles 
& $\xmark$ 
% & 
% 32 fine-grained emotions, 
% 8 empathetic response intents, 
% and 
% the Neutral category
& $-$ 
& $-$ 
& $3,488,300$ utterances
\\ \hline 

EmpathicStories++ 
& $\mathrm{V} + \mathrm{A} + \mathrm{T}$ 
& $\cmark$ 
& $\cmark$ 
& Real-world deployment 
& $\cmark$ 
% & 
% empathy ratings, 6 survey 
% measures (personality, trait 
% absorption, trait empathy, social 
% connection, compassionate love 
% towards humanity, wellbeing)
& 41 
& 53 hrs $/$ 3,180 min 
& 5,380 utterances 
\\  \hline
\end{tabular}
}

    \caption{\textbf{Comparison of \textsc{EmpathicStories++} to related datasets}. In contrast to other datasets, we collect data in-the-wild, over a month-long deployment, and our data is self-annotated with empathy and psychometrics. Since our dataset is interaction-based (we fixed the number of conversation turns per session) and in the real world, we have a limited number of utterances compared to text-only datasets that are crowdsourced from the internet.}
    \label{comparison}
    \vspace{-10pt}

\end{table*}

% \jocelyn{TODO}
\label{sec:relatedwork}
In this work, we present the \textsc{EmpathicStories++} dataset, a multimodal dataset collected from a month-long deployment of social robots in-the-wild. 
In this work, participants shared personal stories with the robot, read stories that were empathically similar to their own experience, and then reflected on stories they empathized with \cite{shen_modeling_2023}. Our interaction design allows researchers to explore empathy in the context of personal experience sharing and understand the influence of users' past experiences on their empathy towards others' experiences. 
We address gaps in previous empathy and emotion recognition datasets through the following attributes: (1) Participant data is captured in their own homes with a social robot. Previous works have shown that users are more comfortable disclosing sensitive information with AI partners than with people \cite{pickard_revealing_2016,lucas_its_2014}. Participants in our study shared emotionally diverse and vulnerable stories from the comfort of their own homes. (2) Participants interacted with the robot over the course of a month, allowing us to obtain longitudinal data. (3) After participants read other peoples' empathically similar stories, they self-rated their empathy towards the story, resulting in more authentic empathy labels. In addition to providing the raw video, audio, and text data for each interaction, we provide extracted features from all three modalities, as well as self-reported psychometric data (i.e. personality, well-being, etc.) and empathy ratings towards other people's stories. These properties enable AI researchers to capture the complexity of empathy in its contextual, longitudinal, and personal dimensions.
In addition to providing the \textsc{EmpathicStories++} dataset, we present a new task on predicting a person's empathy towards other's stories based on their own personal experiences. We evaluate this task in two settings: (1) predicting empathy based on the user's own shared story (e.g. \textit{``I do weekly pickups at the local grocery store to bring leftover food to the food banks...''}), and (2) predicting empathy based on a user's reflection on a story they read (e.g. \textit{``I can really relate to the narrator's feeling of wanting to help others...''}).

\noindent In summary, our contributions are as follows: 
(1) The first multimodal dataset with in-the-wild, long-term, and self-reported cues on empathy towards other people's experiences, containing video, audio, text, as well as low-level features from each modality, self-reported psychometric data, and empathy ratings towards other people's stories. 
(2) A novel task for predicting a user's empathy towards another person's story. (3) Benchmarking empathy prediction using state-of-the-art approaches to enable further improvements in contextual and longitudinal empathy modeling. Our work is a valuable resource for future work in developing social-emotional AI systems, improving interpretability of empathy prediction models, and promoting research on understanding cognitive insights of human empathy.

\section{Related Work} 

% \kt{related works looks very thorough which is great for dataset papers. One thing dumb cvpr reviewers look for is a lot of densely packed citations which you have. }

Relevant prior works span two major areas: (1) social psychological theory on the relationship between prior experience and empathy and (2) datasets containing emotion or empathy ratings used for social-emotional reasoning tasks.

\subsection{Empathy and Memory of Experiences}
Empathy towards others is conditioned on situational (similarity between observer and target) and trait factors (personality, learning history) \cite{davis_empathy_2004, roshanaei_paths_2019}. Furthermore, empathy is tied to important social functions such as prosocial behaviors, social connection, well-being, and psychiatric disorders \cite{morelli_emerging_2015}. A person's past experiences and memories play an important role in both situational and trait empathy. This has been shown clearly in prior work on the social neuroscience of representations of self and other: an observer’s reaction to a target is elicited by language-based cognitive networks that trigger relevant memories with observer’s own feelings \cite{davis_empathy_2004}. Other studies use neuroimaging to show that prosocial behaviors may be due to synchronized representations of self and other \cite{decety_human_nodate}. Memories of other people's past experiences can modulate empathy, as these memories are used to simulate how one might feel in a new situation \cite{ciaramelli_individualized_2013}, and the vividness of memory of others' experiences is tied to prosocial intentions \cite{gaesser_constructing_2013}. 

Besides recalling prior experiences of oneself or others, the process of sharing personal experiences is strongly tied to empathy elicitation. Sharing personal memories makes conversations more truthful, engaging and communicates a person's intentions or feelings \cite{pillemer_remembering_1992, bluck_autobiographical_2003}. The elicited empathy from experience sharing is even stronger when a listener responds with their own personal memories. In empathetic communication, both verbal (vividness of images, verb tense) and nonverbal (emotional gesturing, prosody) cues play a role in perceived empathy \cite{pillemer_remembering_1992, haase_nonverbal_nodate}.

\begin{figure*}[ht]
\centering
         \includegraphics[width=0.75\textwidth]{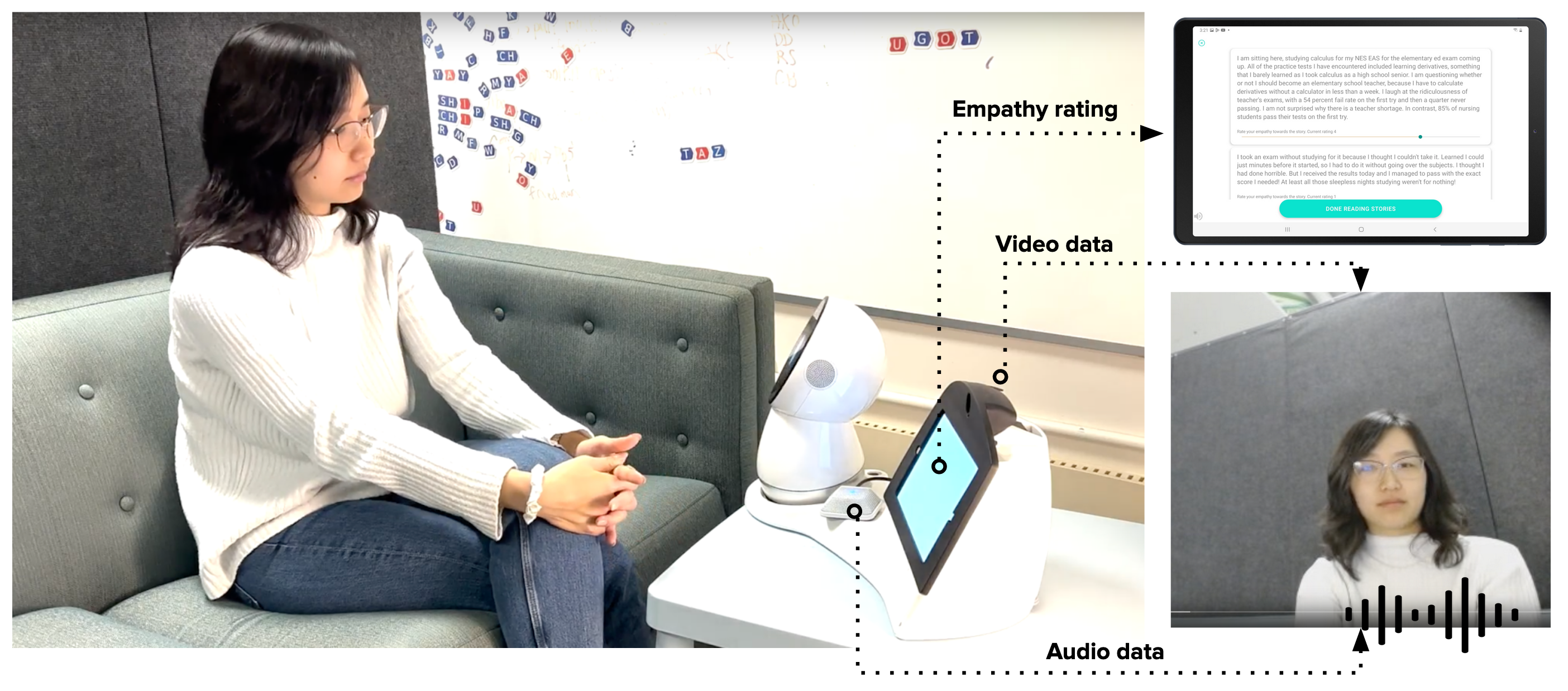}
         \caption{\textbf{Data collection setup. }The robot station houses a webcamera and microphone for video/audio data collection. A tablet displays stories read by participants, as well as sliders for self-rating empathy on a scale of 1-5.}
         \label{station-setup}
         \vspace{-10pt}
\end{figure*}

\begin{figure*}[ht]
  \begin{subfigure}{0.3\textwidth}
      \caption{Video lengths (min)} \label{fig:basicstatsa}
    \includegraphics[width=\linewidth]{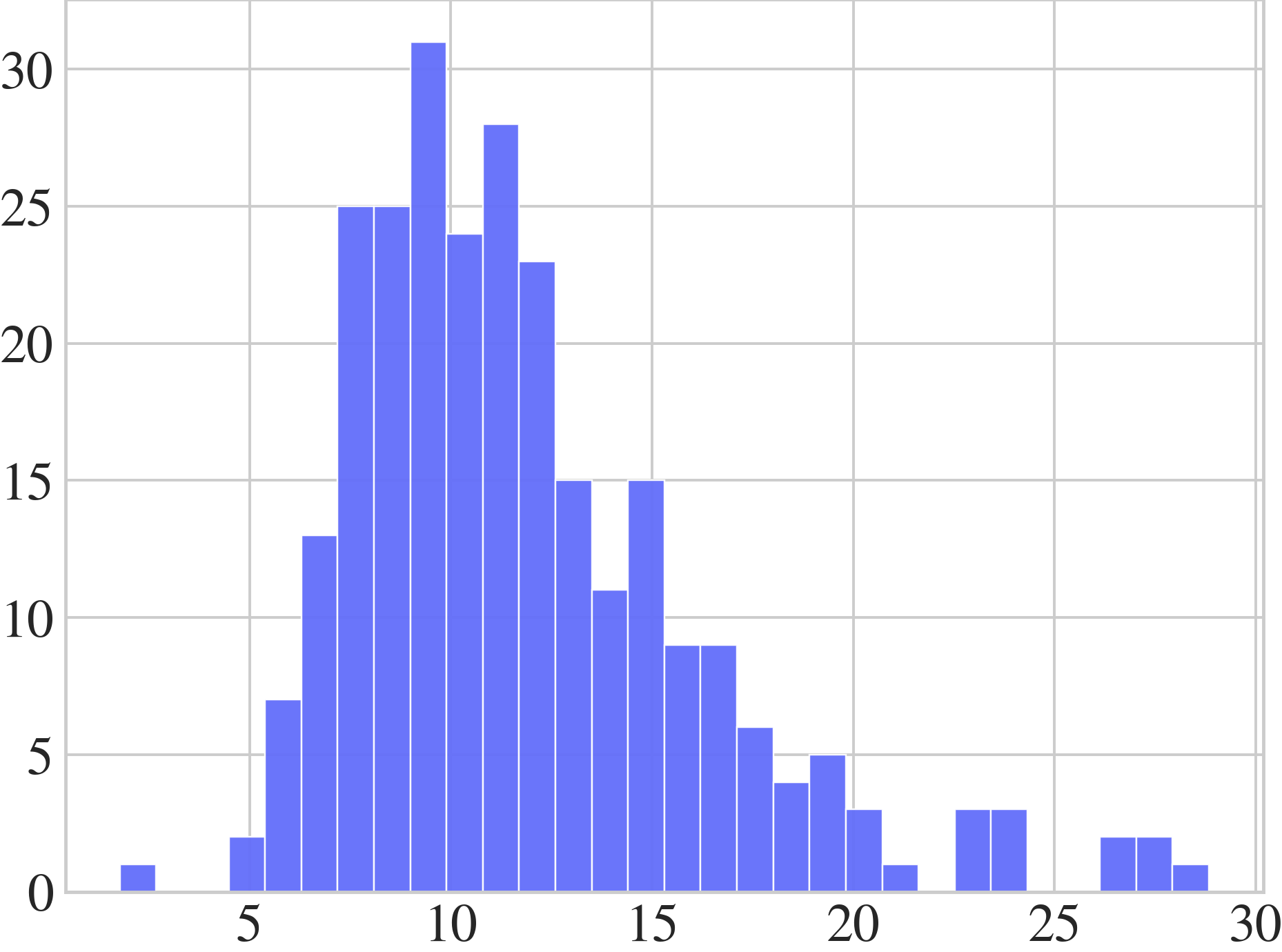}
  \end{subfigure}
  \hfill
  \begin{subfigure}{0.3\textwidth}
    \caption{Words counts} \label{fig:basicstatsb}
    \includegraphics[width=\linewidth]{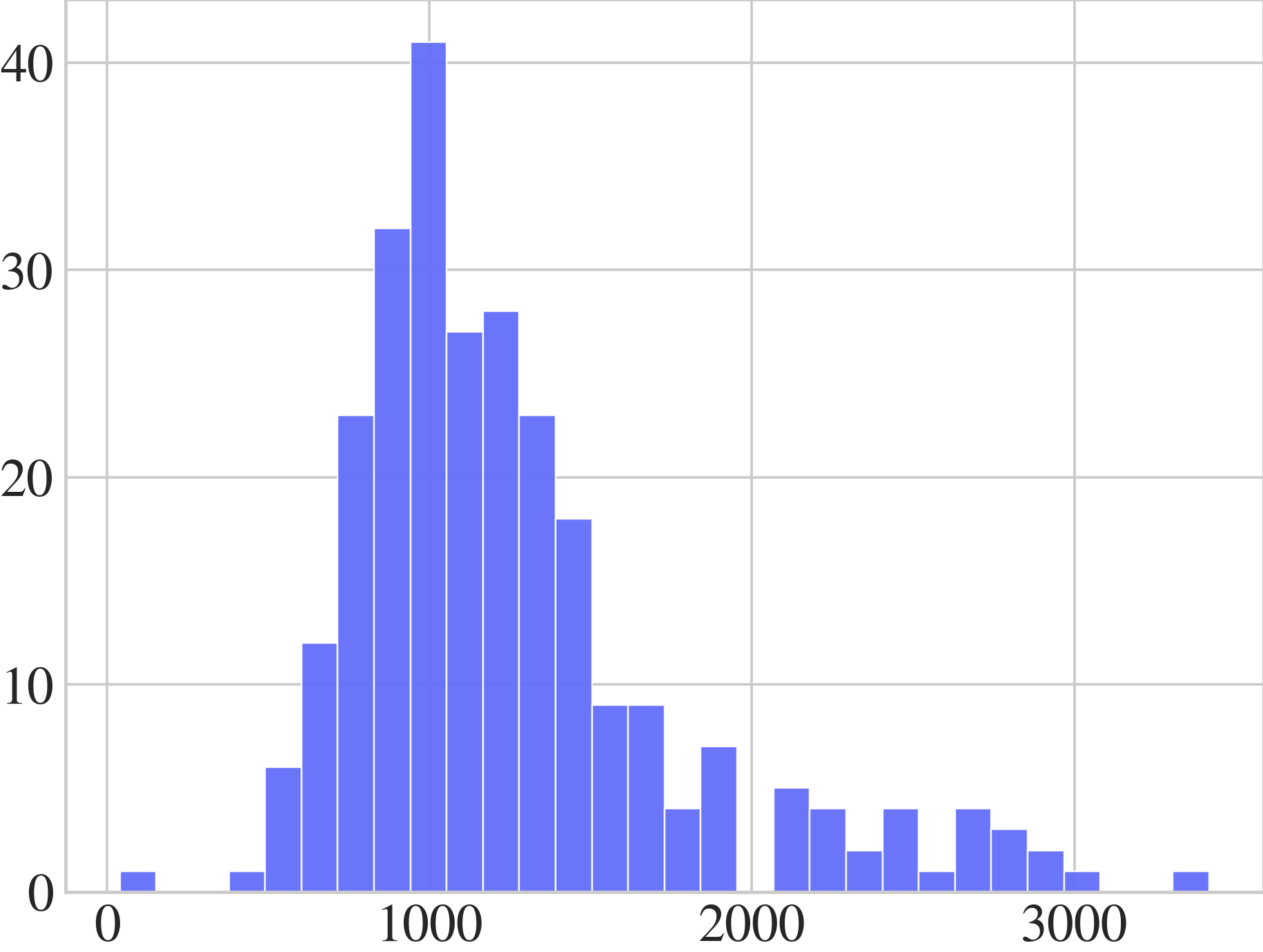}
  \end{subfigure} 
  \hfill
  \begin{subfigure}{0.3\textwidth}
      \caption{Empathy ratings} \label{fig:basicstatsc}
    \includegraphics[width=\linewidth]{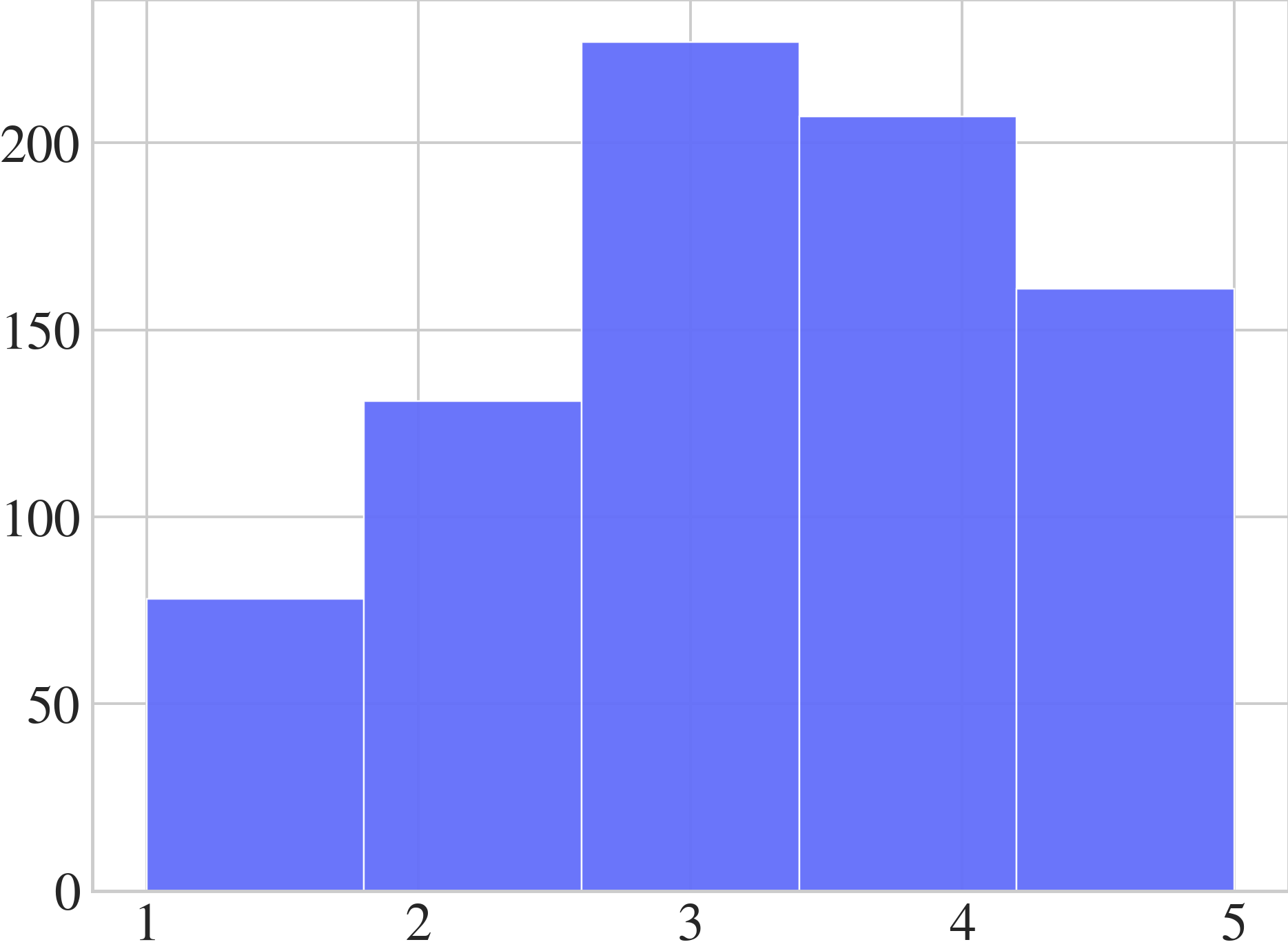}
  \end{subfigure}
  \label{fig:basicstats}
  \caption{\textbf{Basic dataset statistics. }Video length and word count statistics of all participant sessions, as well as the distribution of self-rated empathy labels.}
\vspace{-10pt}

\end{figure*}
% \kt{nice!}
Our dataset addresses all the previous points about empathetic communication: (1) self-reported annotations of situational and trait factors, (2) surveys of relevant social functions including social connection and wellbeing, and (3) video, audio, and transcripts of sessions with participants recalling their own memories and reflecting on others' past experiences over time.

\subsection{Social-Emotional Datasets}
Beyond modeling empathy alone, more broadly, datasets for social and emotional benchmarking have garnered interest in recent years. Datasets such as MELD \cite{poria_meld_2019}, M$^3$ED \cite{zhao_m3ed_2022}, and EmoInt-MD provide multimodal datasets annotated with emotion in conversations pooled from TV shows or movies. The Social-IQ dataset provides a multimodal benchmark for measuring social intelligence \cite{zadeh_social-iq_2019} and the related Social-IQA dataset benchmarks social intelligence with the text modality alone \cite{sap_socialiqa_2019}. There are also datasets that capture the emotions of individuals during story sharing, such as SEND \cite{ong_modeling_2021-1}, emotions of dyads, such as IEMOCAP \cite{busso_iemocap_2008-1} and DAMI-P2C \cite{chen_dyadic_nodate}, as well as datasets of naturalistic conversations, such as the CANDOR dataset \cite{reece_candor_2023}. 

% \kt{nice!}
Few prior works have provided multimodal datasets for empathy tasks alone, and most prior works in empathy benchmarking are text-only. Table \ref{comparison} shows a summary of the most relevant datasets compared to our \textsc{EmpathicStories++} dataset. 
One dataset, the OMG-Empathy dataset measures the emotional effect stories have on the listener \cite{barros_omg-empathy_2019}, but contains a limited amount of data collected from in-lab settings. Two recent works present more substantial datasets: MEDIC, which contains video clips annotated with 3 labels to describe empathy between counselors and clients in psychotherapy sessions \cite{zhouan_zhu_medic_2023}, and a motivational interviewing dataset for assessing therapist empathy \cite{tran_multimodal_2023}. Prior works also provide datasets related to empathy focusing on single modalities. The EmpatheticDialogues dataset \cite{rashkin_towards_2019}, the EDOS dataset \cite{welivita_large-scale_2021}, the EmpathicStories dataset \cite{shen_modeling_2023}, the Empathic Conversations dataset \cite{omitaomu_empathic_2022}, and \citet{sharma_computational_2020} contain text-only benchmarks for empathetic conversations and stories. A few datasets focus on empathy and emotion in nonverbal contexts only, such as the EyeT4Empathy dataset \cite{lencastre_eyet4empathy_2022} and iMiGUE dataset \cite{liu_imigue_2021}, which use gaze and gesture respectively.

% \kt{nice!}
In contrast to these prior works, our dataset is the first dataset that focuses on empathy in relation to past experiences, and is collected in-the-wild, over a long term deployment with longitudinal survey and interaction data, and contains self-annotated empathy ratings.

% Old figure
% \begin{figure*}[ht]
% \centering
%         \includegraphics[width=\textwidth]{figures/basic_stats.png}
%         \caption{Dataset statistics for \textsc{EmpathicStories++} dataset}
%         \label{fig:basicstats}
% \end{figure*}

\section{Data Collection} \label{data-collection}
% \jocelyn{TODO}
\label{sec:datacollection}

We deployed 46 in-home robots, powered by ChatGPT,\footnote{\url{https://chat.openai.com/}} to converse with participants and record data. We recruited participants through mailing lists, and participants explicitly consented to data sharing. Our protocol was approved by our institution's ethics review board. Five participants withdrew from data collection for reasons not related to the study protocol. Data collection took place over the course of a month, and participants were asked to complete between 6-12 conversation sessions with the robot (compensated \$60 for 12 sessions). Figure \ref{station-setup} shows the robot station  in the participants' home and our data collection setup. The use of robots for data collection normalizes speaker-dependent characteristics that could add noise to the data from in-lab, human-human studies or acted scenarios \cite{wood_robot-mediated_2013, wood_robot-mediated_2013-1}. While one might hypothesize that the use of a robot would users less expressive, prior work shows that embodied social agents still elicit empathy behaviors similar to that of human-human interaction \cite{spitale_socially_2022, wood_robot-mediated_2013}. We use the social robot to scaffold the interaction while still allowing for natural conversation.
% Hae Won Park: a good add. any citations to back this up? 
% While this is a nice add to the contribution, we should be prepared to defend the argument of the downsides to using a robot (that users may less express empathy to a robot). For this, we can add citations to support that physically embodied agents elicit empathic response from users.
Within each session, participants were guided through a conversation with the agent using the following scheme. 
% Note that the number of conversation turns in each phase was fixed to control ChatGPT generation:

% \begin{figure*}[ht]
% \centering
%        \includegraphics[width=\textwidth]{figures/combined_stats.png}
%        \caption{Participant demographic information, trait and state survey overviews}
%        \label{fig:combinedstats}
% \end{figure*}

\begin{figure*}[ht]
  \begin{subfigure}[t]{0.26\textwidth}
    \caption{Age} \label{fig:combinedstatsa}
    \smallskip
    \includegraphics[width=\linewidth]{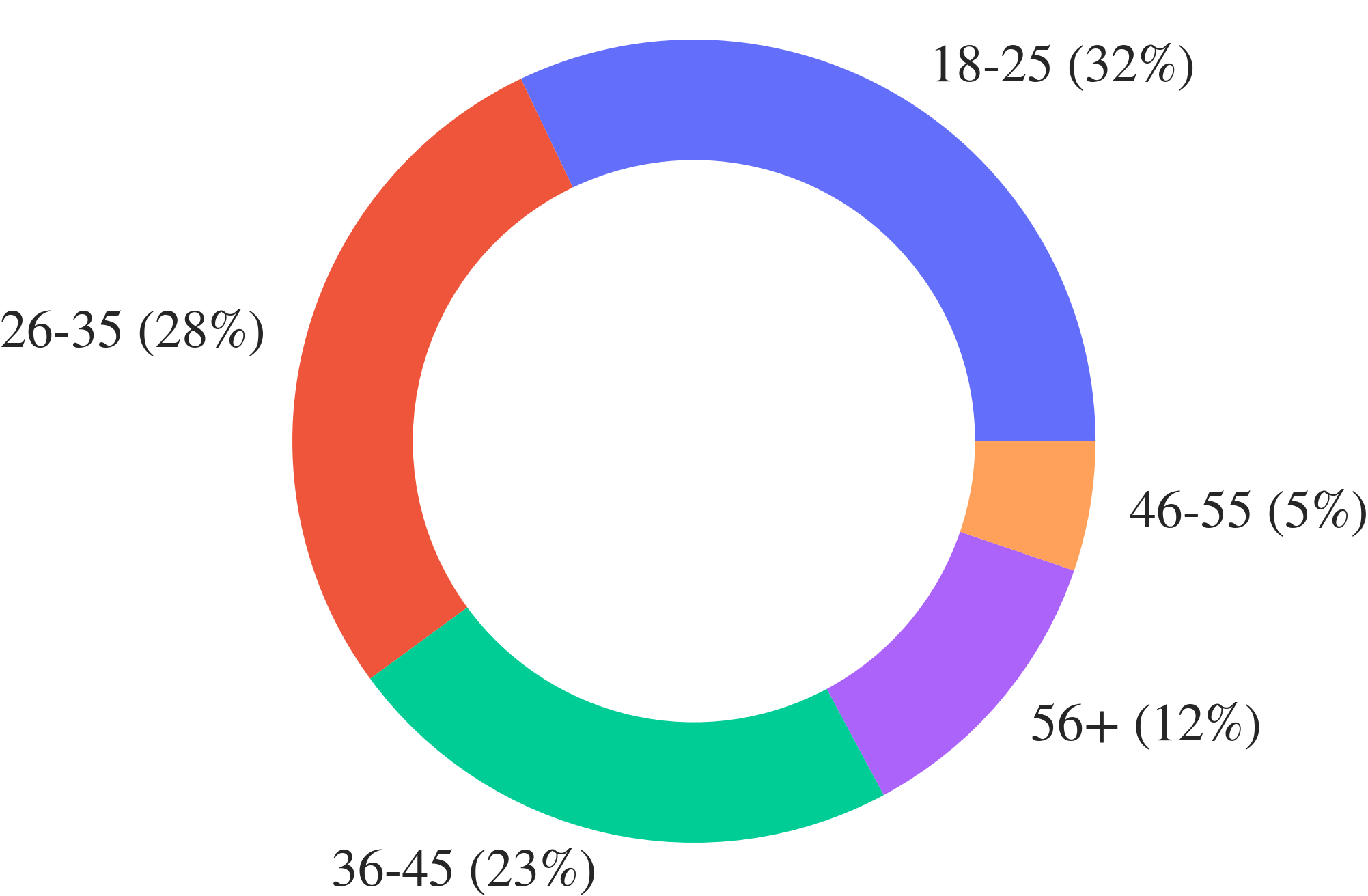}
  \end{subfigure}
  \hfill
  \begin{subfigure}[t]{0.23\textwidth}
    \caption{Gender} \label{fig:combinedstatsb}
    \smallskip
    \includegraphics[width=\linewidth]{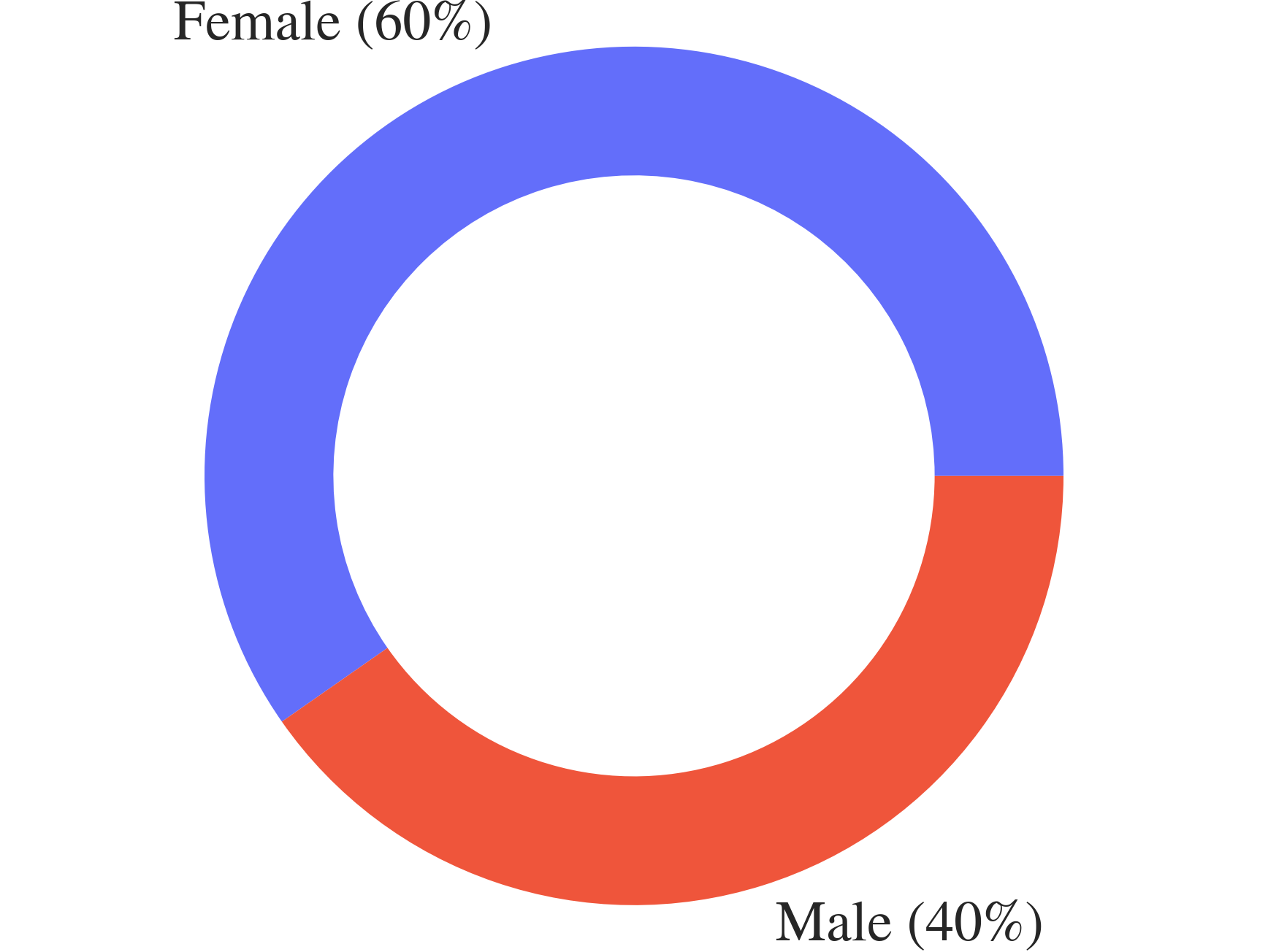}
  \end{subfigure} 
  \hfill
  \begin{subfigure}[t]{0.23\textwidth}
    \caption{Trait empathy} \label{fig:combinedstatsc}
    \smallskip
    \includegraphics[width=\linewidth]{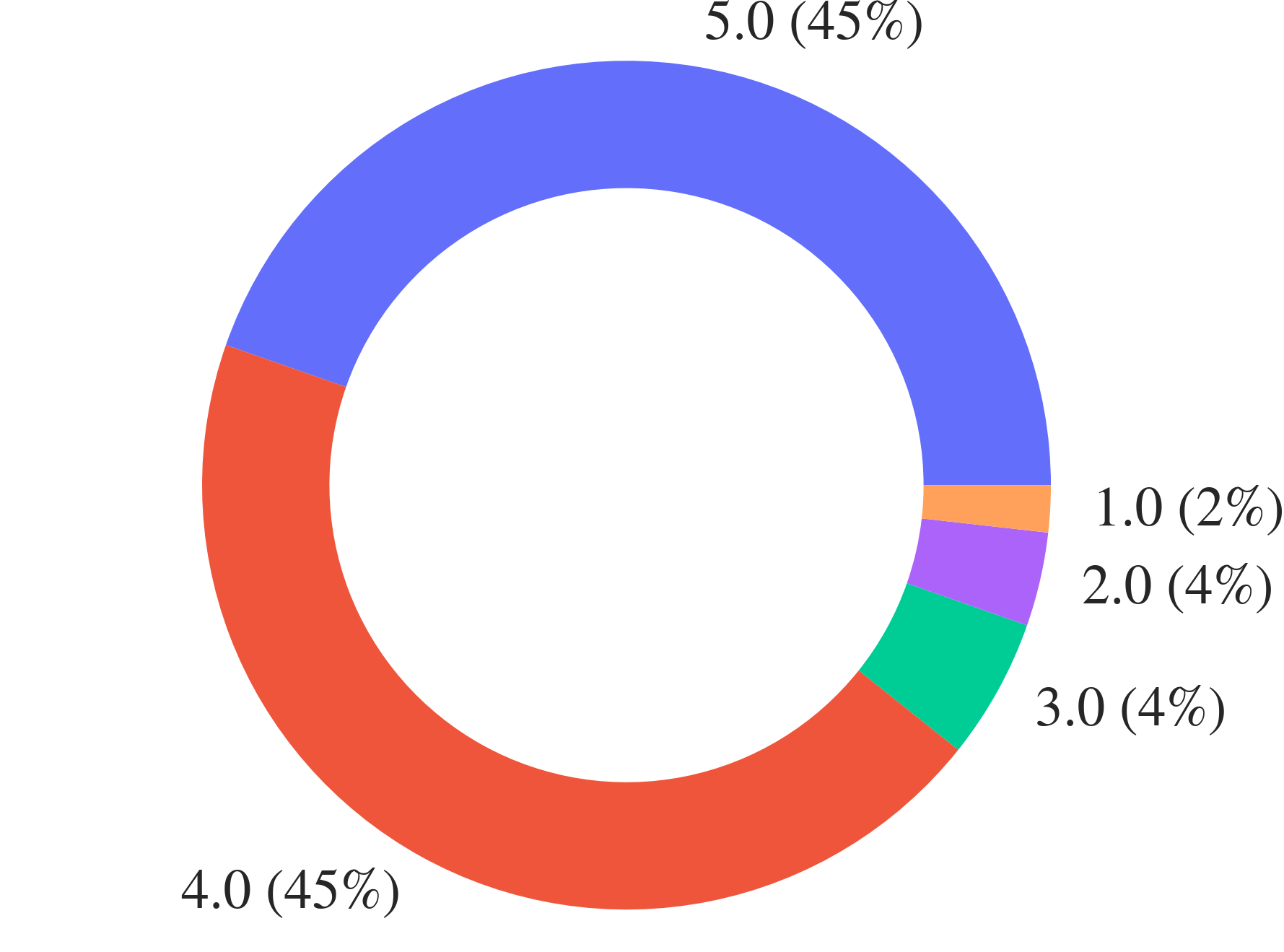}
  \end{subfigure}
  \hfill
  \begin{subfigure}[t]{0.23\textwidth}
    \caption{Trait absorption} \label{fig:combinedstatsd}
    \smallskip
    \includegraphics[width=\linewidth]{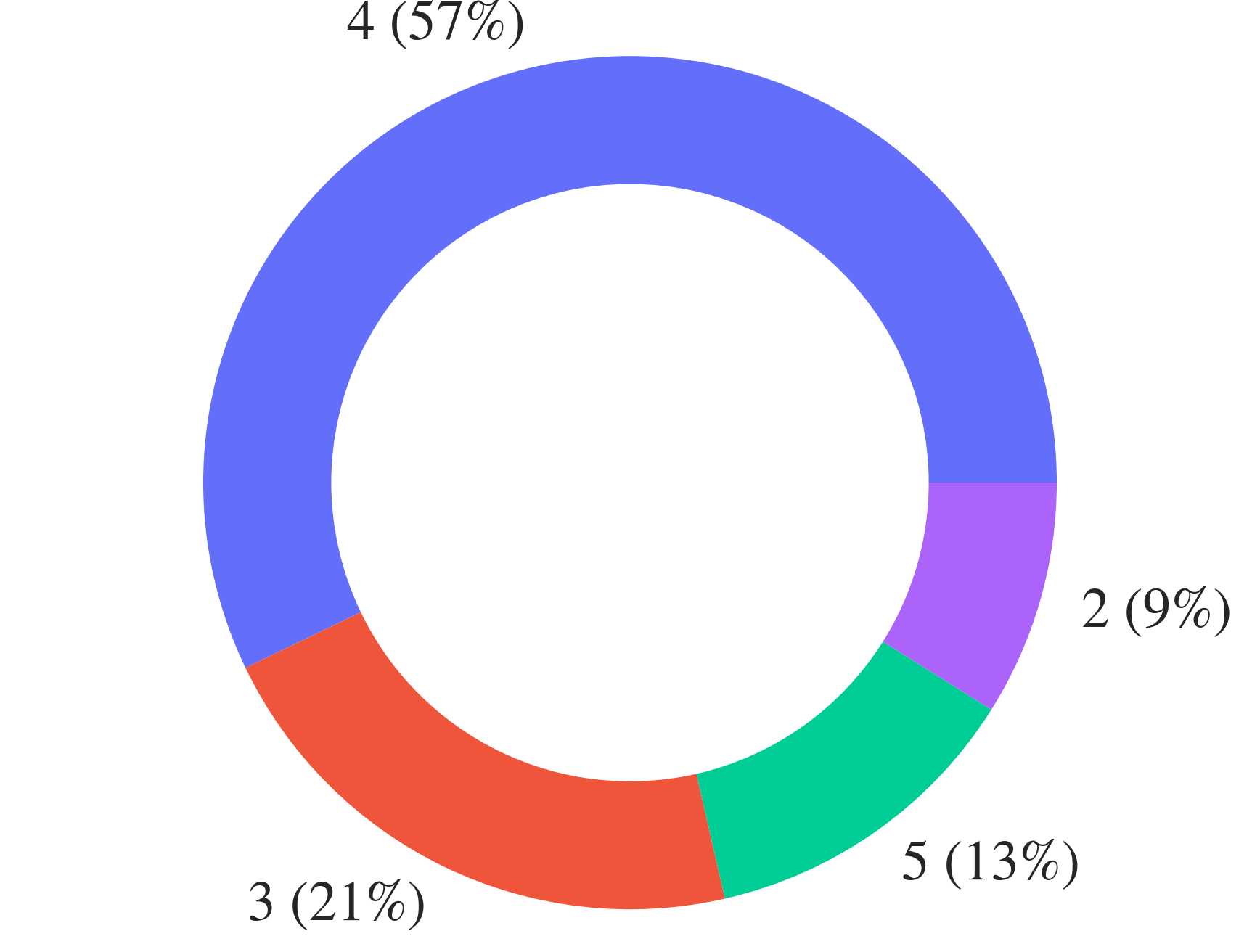}
  \end{subfigure}
  \bigskip

   \begin{subfigure}[t]{0.29\textwidth}
    \caption{Personality} \label{fig:combinedstatse}
    \vspace{0.75cm}
    \includegraphics[width=\linewidth]{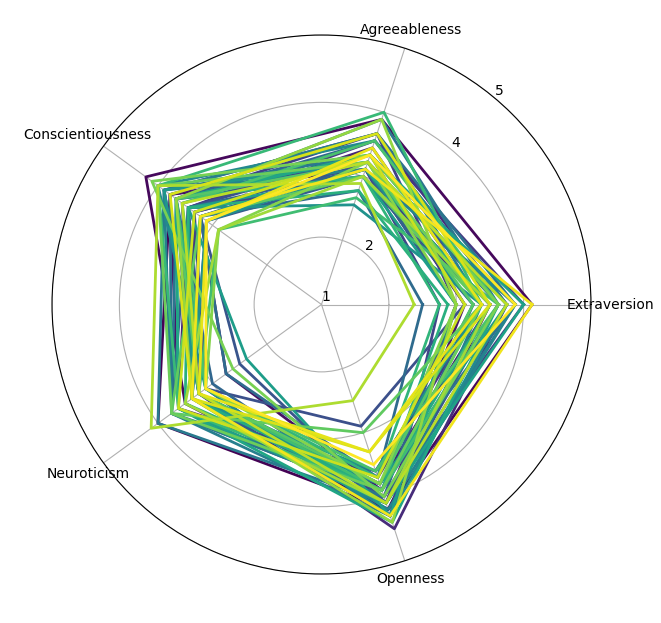}
  \end{subfigure}
  \hfill
  \begin{subfigure}[t]{0.69\textwidth}
    \caption{State surveys} \label{fig:combinedstatsf}
    \smallskip
    \includegraphics[width=\linewidth]{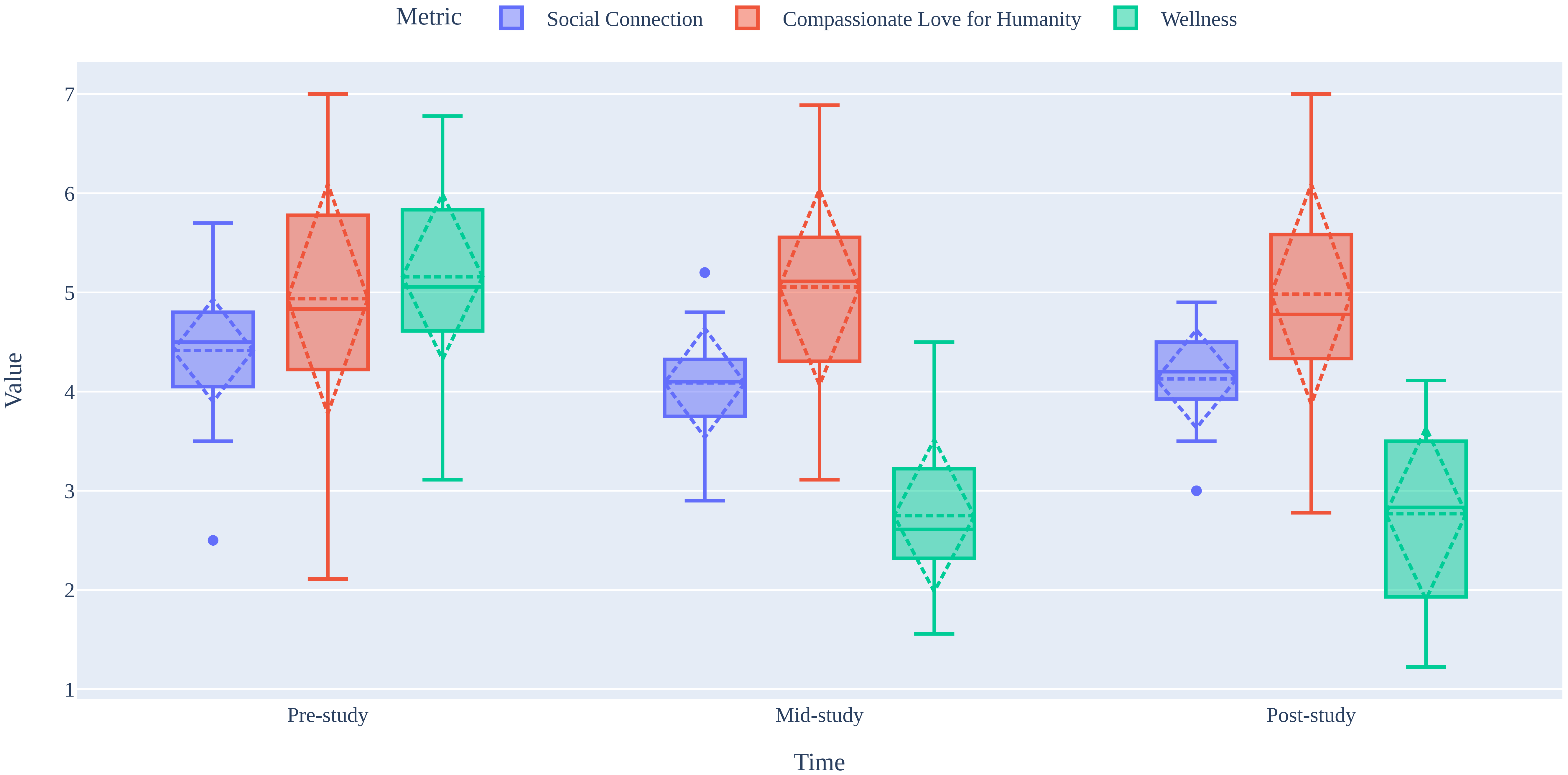}
  \end{subfigure}
  \caption{\textbf{Trait and state surveys}. Participant demographic information, trait and state survey overviews show diversity across age, gender, personality type, and feelings of social connection and wellbeing over time.}
  \label{fig:combinedstats}
           \vspace{-10pt}

\end{figure*}

\begin{enumerate}
    \item \textbf{Warm up phase.} At the beginning of each section, the participant warms up to the robot through casual conversation about their day or the previous robot-participant interaction.

    \item \textbf{Story share phase.} In this phase, the robot prompts the user to share a meaningful story from their journal or on their mind.

    \item \textbf{Story receive phase.} The robot then addresses the user's shared story by responding empathically, and retrieves 3 stories that the user might empathize with, using the empathic similarity retrieval model from \citet{shen_modeling_2023}.
    % On the tablet, the user rates their empathy toward each story using a slider on a scale of 1-5.

    \item \textbf{Story reflection phase.} We carefully designed reflection prompts based on narrative therapy approaches and emotion regulation \cite{gardner_one_2009, white_narrative_1990, yoosefi_looyeh_treating_2014}. Next, the robot asks the participant to reflect on the following four areas: ways in which they related to the narrator, identifying the emotions of the narrator, regulating or comforting the narrator, and high-level takeaways from the story that the participant could apply to their own life.

    \item \textbf{Cool-down phase.} Finally, the agent summarizes the session and thanks the participant.
\end{enumerate}

\paragraph{Self-Report Survey Measures}
We collected self-reported measurements before the study, two weeks into the study, and at the one-month point. During our pre-study questionnaires, we administered the following \textit{trait} surveys:
the Big 5 Personality Test \cite{goldberg_structure_1993}, the absorption scale dimensions of the Multidimensional Personality Questionnaire (measure ability to absorb into fictional experiences)
\cite{noauthor_multidimensional_nodate}, the Single Item Trait Empathy Scale \cite{konrath_development_2018}, and the following \textit{state} surveys: the Compassionate Love for Humanity Scale \cite{sprecher_compassionate_2005}, and the UBC State Social Connection Scale \cite{lok_ubc_2022}. Note that we use both the Compassionate Love for Humanity Scale and the UBC State Social Connection to measure overall ``social connectedness.'' For the mid-study and post-study questionnaires,  all \textit{state} surveys were repeated.
% These surveys focus on different aspects of human connection, with Compassionate Love for Humanity capturing compassion towards \textit{strangers}, whereas the UBC State Social Connection scale captures social connection to both strangers and social partners. 

% \vspace{-10pt}

\paragraph{Interaction Data}
In the \textit{Story receive phase}, participants read three personal stories retrieved based on the user's own story. On the tablet, users rated their empathy toward each story using a slider on a scale of 1-5 (low to high). 

% \vspace{-10pt}

\paragraph{Video and Audio Recordings}
During the study, each station completely recorded each interaction session, using the station's built in Logitech 1080p webcam and MXL AC-44 USB Boundary microphone to obtain high-quality recordings of the participant's face and voice. Note that we made clear when the system was recording the user in our study onboarding and through the robot's ring light. 

% \vspace{-10pt}

\paragraph{Transcripts}
Transcripts of all utterances by the robot and the participant were saved on a Firebase Realtime Database. The transcripts were obtained in real-time using the AssemblyAI streaming ASR.

% TODO: INCLUDE FIGURE/TABLE OF ALL THE DATA WE ARE PROVIDING

% \vspace{-10pt}

\paragraph{Feature Extraction}
For each label, we trimmed the associated video clip to fit a context window of 120 frames ($k = 120$). This gives us 8 seconds worth of video context for videos that play at 15 frames per second. To augment the video clips, we've applied a sliding window technique every second. Consequently, this has yielded us a total of 99,357 clips for the \textit{Story share} phase, and 84,705 samples for the \textit{Reflection} phase.

\begin{itemize}
  \item \textbf{Vision.\footnote{As illustrated in Figure \ref{fig:teaser}, we additionally provide the whole-body (bodies, hands and faces) 2D/3D poses obtained from DOPE \cite{dope} in our dataset.}} We use the normalized eye gaze direction, location of the head, location of 3D landmarks, and facial action units extracted from OpenFace \cite{7477553}. We also extract frame-wise image features from the penultimate layer of ResNet50 \cite{he2015deep}. The two feature vectors (obtained from OpenFace and ResNet50) are concatenated per timestep to be used as the final visual input (dimension/timestep is $F = 2762$).
  \item \textbf{Audio.} We use openSMILE \cite{eyben2010opensmile} to extract low level acoustic features (i.e. loudness, alpha ratio, etc., $F = 65$)
  \item \textbf{Language.} We convert video transcripts and story contents into text embeddings via pre-trained Glove (glove.840B.300d) \cite{pennington-etal-2014-glove} word embedding and Sentence BERT \cite{reimers2019sentencebert}  ($F = 300$, $F = 384$ respectively).
\end{itemize}

\section{Dataset Statistics and Properties}

The \textsc{EmpathicStories++} dataset comprises video, audio, and text data from 269 sessions collected from 41 distinct participants, along with self-reported survey and interaction data. 
Each video is a .avi file recorded at 15fps, whose cumulative length is 3,180 minutes (53 hours). The total number of utterances is 53,80, or about 20 per session (fixed for each interaction phase), totalling 337,147 words (1,258 per session).

\Cref{fig:basicstatsa} shows the distribution of video lengths across sessions, ranging from 2 to 29 minutes (mean = 12 min, s.d. = 4.5 min). \Cref{fig:basicstatsb} depicts a similar distribution for spoken word counts. These ranged between 40 and 3418 words (mean = 1258 words, s.d. = 531 words).
Participants felt varying levels of empathy towards with the stories they received, as the distribution (\Cref{fig:basicstatsc}) of their empathy ratings on 1-5 scale shows (mean = 3.3, s.d. = 1.2).

% Hae Won Park: interesting.. a session length of 2 minutes? is it still a valid session data?
% Nov 16, 2023 12:47 AM

% Hae Won Park: also how did you deal with the sessions without audios for the dataset?

% \kt{nice!}
Figure \ref{fig:combinedstats} depicts the demographic information of the participants. Figures \ref{fig:combinedstatsa}-\ref{fig:combinedstatsd} show the distributions of age, gender, trait empathy, and trait absorption. Measurements of the Big 5 personality traits are shown in the radar chart Figure \ref{fig:combinedstatse}. 
The change in levels of Social Connection, Compassionate Love for Humanity, and Wellness (see Figure \ref{data-collection}) across the month-long study are shown in Figure \ref{fig:combinedstatsf}.
% \vspace{-14pt}
Participants shared vulnerable and meaningful stories across diverse topics (Appendix \ref{appendix:topics}). 
% The main events of each story in each cluster was concatenated and fed to gpt-4 to determine the common topic in them. This common topic becomes the topic of that particular cluster.
% The locations of the topics were then calculated using the centroid of the reduced X and Y coordinates of all the stories belonging to the cluster. The topics are then plotted using the centroid coordinates and remain fixed.
% Then, the stories from each of the participants were embedded using ada-v002. These embeddings are transformed using the UMAP model from the above step to determine the x and y coordinate for it. The story is plotted with those coordinates.
% This method forms a general landscape for journal stories with topic centroids plotted. Then the participant stories are plotted onto the existing landscape to visualize it without being used to create the landscape

% \vspace{-14pt}

% main experiment - story-shared
\renewcommand{\arraystretch}{0.95}% 
\begin{table*}[ht]
\centering
\tiny
\caption{\textbf{Model performance for empathy prediction in \textit{Story Share} scenario} across correlation, accuracy, and retrieval metrics. $r$ = Pearson's correlation, $\rho$ = Spearman's correlation, $Acc$ = Accuracy, $F1$ = Binary F1-score, and $MSE$ = Mean Squared Error. Note that all scores are multiplied by 100 for easier comparison. For each column, the best result is \textbf{bolded}, and the second best is \underline{underlined}.}
\label{tab:model_performance1}
\begin{tabularx}{\textwidth}{l l|X X X X X}
    \hline
    \multicolumn{2}{c|}{Model} & $r$ ($\uparrow$) & $\rho$ ($\uparrow$) & $Acc$ ($\uparrow$) & $F1$ ($\uparrow$) & $MSE$ ($\downarrow$)\\
    \hline
    \multirow{1}{*}{AMER \cite{shen2020memor}} 
    & $t$ & 5.500 \textpm \ 0.800 & 5.500 \textpm \ 0.800 & 53.400 \textpm \ 0.100 & 38.900 \textpm \ 0.600 & 25.200 \textpm \ 0.000 \\ 
    & $v+a$ & 6.300 \textpm \ 0.300 & 6.300 \textpm \ 0.300 & 52.500 \textpm \ 1.600 & 40.000 \textpm \ 0.600 & 25.800 \textpm \ 0.300 \\ 
    & $v+t$ & 4.000 \textpm \ 1.000 & 4.000 \textpm \ 1.000 & 51.800 \textpm \ 0.200 & 38.400 \textpm \ 0.700 & 25.800 \textpm \ 0.100 \\
    & $a+t$ & 6.800 \textpm \ 0.200 & 6.800 \textpm \ 0.200 & 54.100 \textpm \ 0.200 & 39.600 \textpm \ 0.100 & 25.500 \textpm \ 0.000 \\
    & $v+a+t$ & 10.500 \textpm \ 7.000 & 10.500 \textpm \ 7.000 & 51.700 \textpm \ 0.700 & \underline{43.000} \textpm \ 4.900 & 26.400 \textpm \ 0.900 \\
    \hdashline
    TFN \cite{zadeh2017tensor} 
    & $t$ & \underline{11.000} \textpm \ 2.900 & \underline{11.000} \textpm \ 2.900 & 55.100 \textpm \ 1.700 & 41.200 \textpm \ 1.600 & \underline{24.300} \textpm \ 0.100 \\
    & $v+a$ & 0.200 \textpm \ 6.200 & 0.200 \textpm \ 6.200 & 50.700 \textpm \ 2.800 & 34.600 \textpm \ 3.700 & 24.400 \textpm \ 0.200 \\
    & $v+t$ & -4.700 \textpm \ 10.900 & -4.700 \textpm \ 10.900 & 48.100 \textpm \ 4.900 & 32.000 \textpm \ 6.300 & 24.400 \textpm \ 0.500 \\
    & $a+t$ & -4.400 \textpm \ 9.300 & -4.400 \textpm \ 9.300 & 48.600 \textpm \ 4.500 & 32.100 \textpm \ 5.200 & 24.400 \textpm \ 0.200 \\
    & $v+a+t$ & -1.900 \textpm \ 9.300 & -1.900 \textpm \ 9.300 & 50.200 \textpm \ 3.000 & 33.100 \textpm \ 6.300 & \textbf{24.200} \textpm \ 1.000 \\
    \hdashline
    EF-LSTM \cite{hochreiter1997long} 
    & $t$ & 3.000 \textpm \ 2.300 & 3.000 \textpm \ 2.300 & 52.500 \textpm \ 1.000 & 37.300 \textpm \ 1.400 & 25.300 \textpm \ 0.200 \\
    & $v+a$ & 4.800 \textpm \ 3.400 & 4.800 \textpm \ 3.400 & 51.300 \textpm \ 0.700 & 39.300 \textpm \ 2.300 & 26.000 \textpm \ 0.200 \\
    & $v+t$ & 7.900 \textpm \ 1.300 & 7.900 \textpm \ 1.300 & 50.200 \textpm \ 2.000 & 42.000 \textpm \ 1.300 & 26.600 \textpm \ 0.700 \\
    & $a+t$ & 2.800 \textpm \ 2.100 & 2.800 \textpm \ 2.100 & 52.300 \textpm \ 0.700 & 37.200 \textpm \ 1.400 & 25.400 \textpm \ 0.100 \\
    & $v+a+t$ & 7.400 \textpm \ 1.000 & 7.400 \textpm \ 1.000 & 51.200 \textpm \ 2.700 & 41.400 \textpm \ 0.400 & 26.300 \textpm \ 0.700 \\
    \hdashline
    LF-LSTM \cite{hochreiter1997long} 
    & $t$ & 3.400 \textpm \ 0.100 & 3.400 \textpm \ 0.100 & 52.600 \textpm \ 0.100 & 37.600 \textpm \ 0.000 & 25.100 \textpm \ 0.000 \\
    & $v+a$ & 6.400 \textpm \ 1.600 & 6.400 \textpm \ 1.600 & 46.000 \textpm \ 0.600 & 42.500 \textpm \ 0.900 & 27.800 \textpm \ 0.200 \\
    & $v+t$ & 5.800 \textpm \ 4.600 & 5.800 \textpm \ 4.600 & 47.200 \textpm \ 2.800 & 41.800 \textpm \ 2.000 & 27.600 \textpm \ 0.600 \\
    & $a+t$ & 2.200 \textpm \ 1.300 & 2.200 \textpm \ 1.300 & 52.300 \textpm \ 0.700 & 36.700 \textpm \ 0.800 & 25.300 \textpm \ 0.000 \\
    & $v+a+t$ & 8.200 \textpm \ 5.000 & 8.200 \textpm \ 5.000 & 48.100 \textpm \ 0.800 & 42.800 \textpm \ 3.300 & 27.200 \textpm \ 0.800 \\
    \hdashline
    EmpathicStoriesBART \cite{shen_modeling_2023} 
    & $t$ & 2.400 \textpm \ 0.000 & 2.400 \textpm \ 0.000 & \underline{80.700} \textpm \ 0.000 & 35.500 \textpm \ 0.000 & 51.900 \textpm \ 0.000 \\
    \hdashline
    GPT-4 \cite{openai2023gpt4} 
    & $t$ & \textbf{23.200} \textpm \ 1.600 & \textbf{17.600} \textpm \ 1.400 & \textbf{82.500} \textpm \ 0.000 & \textbf{50.600} \textpm \ 0.700 & 32.200 \textpm \ 0.300 \\
    % add more rows as needed
    \hline
\end{tabularx}
\vspace{-10pt}
\end{table*}

Our dataset is notable in that it (1) \textbf{is captured in-the-wild}, in participants' homes (2) \textbf{contains longitudinal data, with trait and state surveys}, and (3) \textbf{is self-annotated}, which is crucial for a subjective psychological process like empathy. 
% In this section, we summarize the video, audio, and text properties as well as the participant statistics,  trait and state survey statistics, self-annotated empathy rating statistics.

% DATA EXPLORATION IDEAS:
%   top correlations according to LWIC (thse are weird though)
%   sentiment analysis and emotion classification distributions (from HuggingFace?)
%       on shared, received, selected, unselected, reflection
%   for sentiment analysis compare correlations of 
%       shared, selected (if positive shared then likely to select positive?),
%       shared, unselected (if negative shared then less likely to select positive?),
%       selected, reflection (does reflection depend more on the selected story
%       shared, reflection      or the shared one?)

\section{Experiments}\label{sec:models}

\subsection{Task Definition} 
% \yubin{TODO}
% We formulate the multimodal empathic prediction problem as the following. We are given video clips of in SITU. We trim the videos into clips of $k$ frames within window size of $w$ centering to the annotation timestamp. At time $t$, where $t$ is the timestep in which we want to predict each participant’s empathy levels for three stories, we are given the $[t-k, ..., t]$ interval of contextual video information; $k$ is the number of frames we use as context. For each clip $p$, their corresponding contextual behavior features can be viewed as $X = []$. We train a model that takes $X$ as input and predict the empathy level for three stories $\hat{y} \in (0,1)^{C}$ 

We formulate the multimodal empathy prediction task as follows: At time $t$, where $t$ is the timestep in which we want to predict each participant’s empathy levels for the story, we are given the $[t-k/2, ..., t+k/2]$ interval of contextual video information (during the \textit{Story Share} and \textit{Reflection} phases), where $k$ is the number of context frames. 

For each clip, we extract features from three modalities: text, audio, and video. Each modality has distinct temporal and feature dimension, denoted as $T_{\{V,A,T\}} \times F_{\{V,A,T\}}$. The corresponding contextual behavior features for each modality can be viewed as $X_{T} \in \mathbb{R}^{T_{T} \times F_{T}}$, $X_{A} \in \mathbb{R}^{T_{A} \times F_{A}}$, and $X_{V} \in \mathbb{R}^{T_{V} \times F_{V}}$, respectively. The comprehensive multimodal feature set is represented as $X = [X_{T}, X_{A}, X_{V}]$. Finally, we train a model $f_{\theta}(\cdot)$ that takes $X$ as input and outputs a multimodal representation $Z = f_{\theta}(X)$, which is further used to calculate empathic similarity score $sim(Z, E(S_{i}))$ where $sim(\cdot)$ is a similarity metric (e.g., cosine similarity), and $E(S_i)$ is the embedding of the $i$th story $S_i$ ($i = 1,2,3$). Finally, this similarity score is compared with the empathic label $y$ to calculate the loss.

\subsection{Models} 
% \yubin{TODO}
\paragraph{Attention-based multimodal Emotion Reasoning model (AMER) \cite{shen2020memor}:} AMER is a model designed to facilitate the task of multimodal emotion reasoning in videos. It employs an attention-based approach to model intra- and inter- personal emotion contexts, propagation, and prior knowledge of personalities. 
% AMER encodes multimodal features into compact representations, recognizing the emotion of a target person in a given video segment. 

% \paragraph{Memory Fusion Network (MFN) \cite{zadeh2018memory}:} The MFN model is adept at tasks such as sentiment analysis and emotion recognition. One of its key features is the use of a delta-memory attention mechanism that tracks the sequential changes across multiple LSTM layers. This model also aligns and fuses information from transcript, audio, and video modalities, storing the resulting multimodal insights in a separate memory unit. This alignment process is crucial for a comprehensive understanding of multimodal language data.

% \vspace{-12pt}

\paragraph{Tensor Fusion Network (TFN) \cite{zadeh2017tensor}:} TFN is a representative tensor-based network, initially developed for multimodal sentiment analysis. It carries out an outer tensor-product operation on the embeddings of modalities to create a unified multimodal space. 
% It stands out for its ability to perform fusion operations on unimodal, bimodal, and trimodal data components. The fusion output is then integrated with the question and answer data to make the final prediction.

% \vspace{-12pt}

\paragraph{Late-Fusion LSTM (LF-LSTM) \cite{hochreiter1997long}:} LF-LSTM is a model that separately constructs LSTMs for linguistic, visual, and acoustic inputs. It fuses the final hidden states of these three LSTMs, creating a comprehensive sentence-level multimodal representation. 
% This representation is then sequentially transmitted through two fully-connected layers to generate the final output, showcasing its capabilities in multimodal sentiment analysis tasks.

% \vspace{-12pt}

\paragraph{Early-Fusion LSTM (EF-LSTM) \cite{hochreiter1997long}:} EF-LSTM assembles linguistic, visual, and acoustic features at each time step, utilizing an LSTM to construct a sentence-level multimodal representation. 
% The final hidden state is then extracted and sequentially forwarded through two fully-connected layers to yield the sentiment output.

% \paragraph{Sentimental Words Aware Fusion Network (SWAFN) \cite{chen2020swafn}:} SWAFN is a multimodal sentiment analysis tool that integrates language, vision, and acoustic features. It employs a cross-modal co-attention mechanism for context understanding and a multi-task approach focused on sentimental words classification to guide the fusion of modalities. Through this method, SWAFN effectively incorporates external language knowledge, enhancing the learning of powerful multimodal representations.

% \vspace{-12pt}

\paragraph{EmpathicStoriesBART \cite{shen2023modeling}:} EmpathicStoriesBART is a distinctive model fine-tuned to compute empathic similarity in personal narratives using three key story features. Validated in a user study, it outperforms traditional semantic similarity models, highlighting its potential for our task.

% \vspace{-12pt}

\paragraph{GPT-4 \cite{openai2023gpt4}:} GPT-4, a state-of-the-art closed-source language model  capable of deep contextual understanding and producing highly relevant responses. GPT models have been evaluated for empathetic response generation \cite{lee_does_nodate}. 

Implementation details and prompts are included in Appendix \ref{implementation} and \ref{prompts}.

% main experiment - reflection

\section{Results and Discussion}

\subsection{Automatic Evaluation}
To evaluate the quality of empathy predictions, we follow previous work \cite{shen_modeling_2023} and report Pearson’s correlation, Spearman’s correlation, accuracy, F1-scores and the mean squared error. For correlations, we calculate the cosine similarity between the multimodal representation and the embedding of the stories and compare these similarity scores with the human-rated empathy labels. For interpretability, we split the scores into binary similar/dissimilar categories and compute the accuracy and $F1$ scores.

Table \ref{tab:model_performance1} shows the performance of state-of-the-art multimodal (video, audio and text) models when given the user's \textit{Story Share} context (video and audio) + the story they read (text) as inputs, and their empathy ratings as labels. 
% These tables represent the participants' evaluation of their empathy levels with the three given stories, read alongside robots in a video dataset, according to their personal stories. Subsequently, during the \textit{Reflection} phase, they expressed their sentiments about a single story. 
In the context of \textit{Story Share}, GPT-4 showed the highest Pearson's correlation ($r$ = 0.232) and Spearman's correlation ($\rho$ = 0.176) with $t$-only input. Notably, it also recorded the highest accuracy ($Acc = 0.825$) and $F1$-score ($F1$ = 0.506) which aligns well with the observation that participants in the \textit{Story Share} setting were more focused on conveying their story, rather than on expressive verbal and non-verbal behaviors.
% Conversely, TFN exhibited the best results with $a+t$ inputs. Meanwhile, EF-LSTM, which constructs a sentence-level multimodal representation, showed the highest accuracy ($acc$ = 0.556) with $v$+$a$ modal inputs. 
Conversly, the performance of models in the context of user \textit{Reflection} (reflections on a read story) is outlined in Table \ref{tab:model_performance2}. Here, LF-LSTM demonstrated the highest Pearson's correlation ($r$ = 0.560) and Spearman's correlation ($\rho$ = 0.559) with $v$+$t$ inputs. While GPT-4 continued to show the highest accuracy and F1-scores, it's worth noting that among multimodal models, AMER showed comparable performance ($Acc = 0.688$, $F1 = 0.665$) even with a significantly smaller number of parameters and using only audio with text inputs.
% On the other hand, the TFN model showed the best performance with $t$ inputs, while the EF-LSTM model exhibited the highest accuracy ($acc$ = 0.592) with $v$+$a$ modal inputs.

\renewcommand{\arraystretch}{0.95}% 
\begin{table*}[ht!]
\centering
\tiny
\caption{\textbf{Model performance for empathy prediction in the \textit{Reflection} scenario.} For each column, the best result is \textbf{bolded}, and the second best is \underline{underlined}.}
\label{tab:model_performance2}
\begin{tabularx}{\textwidth}{l l|X X X X X}
\hline
\multicolumn{2}{c|}{Model} & $r$ ($\uparrow$) & $\rho$ ($\uparrow$) & $Acc$ ($\uparrow$) & $F1$ ($\uparrow$) & $MSE$ ($\downarrow$)\\
\hline
\multirow{1}{*}{AMER \cite{shen2020memor}} 
& $t$ & 5.400 \textpm \ 0.700 & 5.300 \textpm \ 0.700 & 53.900 \textpm \ 1.400 & 43.500 \textpm \ 1.600 & 23.800 \textpm \ 0.700 \\
& $v+a$ & 36.500 \textpm \ 0.500 & 36.600 \textpm \ 0.500 & 68.400 \textpm \ 0.100 & 65.400 \textpm \ 0.500 & \underline{22.500} \textpm \ 0.000 \\
& $v+t$ & 40.000 \textpm \ 0.400 & 39.900 \textpm \ 0.400 & 67.800 \textpm \ 0.200 & 66.300 \textpm \ 6.600 & \textbf{22.400} \textpm \ 0.000 \\
& $a+t$ & 37.300 \textpm \ 0.200 & 37.200 \textpm \ 0.100 & \underline{68.800} \textpm \ 0.000 & 66.500 \textpm \ 0.100 & \underline{22.500} \textpm \ 0.000 \\
& $v+a+t$ & 36.700 \textpm \ 0.500 & 36.800 \textpm \ 0.600 & 68.400 \textpm \ 0.400 & 65.400 \textpm \ 1.100 & \underline{22.500} \textpm \ 0.100 \\
\hdashline
TFN \cite{zadeh2017tensor}
& $t$ & 0.500 \textpm \ 4.800 & 0.500 \textpm \ 4.200 & 51.100 \textpm \ 2.600 & 32.200 \textpm \ 3.200 & 23.700 \textpm \ 0.100 \\
& $v+a$ & 2.300 \textpm \ 4.600 & 2.400 \textpm \ 4.500 & 49.800 \textpm \ 1.100 & 32.600 \textpm \ 2.000 & 23.800 \textpm \ 0.300 \\
& $v+t$ & -0.100 \textpm \ 5.500 & 0.000 \textpm \ 5.400 & 52.000 \textpm \ 3.300 & 32.700 \textpm \ 4.700 & 23.600 \textpm \ 0.200 \\
& $a+t$ & 5.100 \textpm \ 4.600 & 5.100 \textpm \ 4.600 & 51.500 \textpm \ 1.400 & 30.000 \textpm \ 1.300 & 23.800 \textpm \ 0.200 \\
& $v+a+t$ & 8.400 \textpm \ 4.000 & 8.300 \textpm \ 4.000 & 49.400 \textpm \ 0.800 & 32.800 \textpm \ 0.400 & 23.900 \textpm \ 0.200 \\
\hdashline
EF-LSTM \cite{hochreiter1997long}
& $t$ & 6.000 \textpm \ 0.700 & 5.900 \textpm \ 0.700 & 53.500 \textpm \ 0.200 & 30.000 \textpm \ 0.600 & 23.400 \textpm \ 0.100 \\
& $v+a$ & 24.500 \textpm \ 1.400 & 24.500 \textpm \ 1.400 & 61.100 \textpm \ 1.400 & 44.900 \textpm \ 0.900 & 23.400 \textpm \ 0.600 \\
& $v+t$ & 22.600 \textpm \ 4.300 & 22.400 \textpm \ 3.400 & 58.700 \textpm \ 3.500 & 43.900 \textpm \ 2.400 & 24.100 \textpm \ 1.100 \\
& $a+t$ & 2.500 \textpm \ 4.200 & 2.500 \textpm \ 4.200 & 54.900 \textpm \ 1.500 & 31.000 \textpm \ 2.700 & 23.400 \textpm \ 0.000 \\
& $v+a+t$ & 20.500 \textpm 4.100 & 20.500 \textpm 4.100 & 57.300 \textpm 3.900 & 42.800 \textpm 2.200 & 24.300 \textpm 1.000 \\
\hdashline
LF-LSTM \cite{hochreiter1997long}
& $t$ & 3.100 \textpm \ 0.500 & 3.100 \textpm \ 0.500 & 55.400 \textpm \ 0.400 & 31.200 \textpm \ 1.200 & 23.200 \textpm \ 0.000 \\
& $v+a$ & 1.900 \textpm \ 5.300 & 1.900 \textpm \ 5.300 & 54.100 \textpm \ 2.000 & 30.900 \textpm \ 3.300 & 23.500 \textpm \ 0.300 \\
& $v+t$ & \textbf{56.000} \textpm \ 4.900 & \textbf{55.900} \textpm \ 4.900 & 56.700 \textpm \ 1.700 & 32.700 \textpm \ 3.200 & 23.100 \textpm \ 0.100 \\
& $a+t$ & 2.400 \textpm \ 0.500 & 2.400 \textpm \ 0.500 & 54.900 \textpm \ 0.700 & 30.900 \textpm \ 0.100 & 23.300 \textpm \ 0.100 \\
& $v+a+t$ & 6.800 \textpm 8.700 & 6.800 \textpm 8.700 & 57.200 \textpm 3.900 & 33.600 \textpm 5.500 & 23.000 \textpm 0.200 \\
\hdashline
EmpathicStoriesBART \cite{shen_modeling_2023}
& $t$ & 32.700 \textpm \ 0.000 & 34.000 \textpm \ 0.000 & \textbf{73.700} \textpm \ 0.000 & \textbf{76.200} \textpm \ 0.000 & 34.900 \textpm \ 0.000 \\
\hdashline
GPT-4 \cite{openai2023gpt4}
& $t$ & \underline{50.800} \textpm \ 0.000 & \underline{46.000} \textpm \ 0.000 & \textbf{73.700} \textpm \ 0.000 & \underline{75.000} \textpm \ 0.000 & 30.000 \textpm \ 0.000 \\
\hline
\end{tabularx}
\vspace{-10pt}
\end{table*}

\subsection{Ablation Studies}
Here, we analyze the influence of various input modalities on six models in both \textit{Story Share} and \textit{Reflection} settings, focusing particularly on the impact of text-only inputs.

In the \textit{Story Share} scenario, across different models and input modalities, no significant performance improvements were observed as we add more input modalities to text. Interestingly, using $t$-only input showed the best performance in $Acc$ across all multimodal models. In contrast, in the \textit{Reflection} scenario, where both verbal and non-verbal expressions plays a vital role, AMER showed remarkable performance improvements (26.90\% in $Acc$) when adding $a$ to $t$ and 14.02\% for EF-LSTM when using $v$+$a$ inputs. Also, by adding $v$ to $t$, all multimodal model showed performance improvements (10.36\% for $Acc$ and 26.28\% for $F1$ in average). However, EmpathicStoriesBART and GPT-4 model, which solely use $t$-only input, outperforms all other models, achieving an impressive accuracy of 0.737. This significant performance, combined with a high $F1$ score of 0.762 and 0.750, underscores the potential of task and context specificity and the use of key story features to identify moments of empathy. To confirm the robustness of these results, we applied majority voting based on the label distribution, which resulted in an accuracy of 0.750 and an $F1$ score of 0.839.

% Overall, the selection of input modalities considerably impacts the performance of empathic similarity prediction tasks. While multimodal inputs can enhance performance in scenarios like reflection, where non-verbal expressions are abundant while people tell their own experience, transformer-based models like GPT-4 and EmpathicStoriesBART demonstrate their proficiency in reflection scenarios dominated by introspective verbal expressions. This provides insights into the alignment of model choice and input modalities with the specific characteristics of the scenario in focus, highlighting the potential for multimodal models in social-emotional AI.

% mhulse: I think the **only** reason these models seem to perform this well is because: (a) participants only reflected on stories they rated highly (b) accuracy and F1 was computed for these models by determining an optimal threshold similarity to declare high empathy; because of (a), the  threshold in (b) is extremely low (i.e. just say all inputs are highly empathetic). This renders the accuracy score basically meaningless.

% tl;dr: a model that says "high empathy" for every input would perform just as well under these evaluation metrics.

% \subsection{Interpretability Analysis}  \yubin{TODO}
% \subsection{Insights on Empathy Behaviors} \jocelyn{TODO}

\section{Conclusion} 
This paper presents \textsc{EmpathicStories++}, the first in-the-wild, long-term, multimodal dataset on empathy towards personal experiences, which can be used to quantitatively evaluate empathy as it relates to one's past experiences. 
Our dataset is self-annotated with empathy ratings and psychometric surveys. 
We present and benchmark a task on predicting user empathy from their interaction contexts. 
We observe that modality selection impacts model performance and is context-dependent. In the \textit{Story Share} phase, where textual context was dominant, GPT-4 with text input performed the best in most metrics. However, in the \textit{Reflection} phase, where introspective verbal and non-verbal expressions are abundant, using $v$+$t$ inputs showed 26.28\% improvement in average for $F1$ score, demonstrating their proficiency in extracting meaningful insights from multi-modal inputs. 
Our work provides a valuable resource for future work in empathetic AI, quantitative exploration of cognitive insights, and empathy modeling. We publicly release our dataset to foster advancements in social-emotional AI.

\section*{Acknowledgments}
We would like to thank the participants of our work for contributing to this dataset. We would also like to thank Jon Ferguson and Audrey Lee for their technical contributions in deploying the robot stations for data collection. This work was supported by an NSF GRFP under Grant No. 2141064

\section*{Ethics Statement}
Our dataset contains intimate, personal stories and video-audio data of participants necessary for modeling empathetic response. However, this type of naturalistic data is sensitive and private. As such, we made sure participants explicitly consented with data sharing, and our protocol was approved by our institutions ethics review board. Furthermore, in the design of our robot station, we made sure it was clear whenever the robot was listening (through a blue ring light) and that data would only be recorded during sessions, not the entire duration the robot was in a participant's home. We made sure to store videos on a private, lab-hosted server. For transparency we note that 17\% of participants mentioned concerns of the robot infringing their privacy/security during our post-study interviews. While our dataset contains intimate information, we believe that such a resource is necessary in advancing science about empathy, which by nature occurs in personal and natural settings. We will ensure that distribution of the dataset is only granted upon ethics review board approval, and that the dataset is only used towards the goal of furthering positive empathy research in the future.

% Scientific work published at ACL 2023 must comply with the ACL Ethics Policy.\footnote{\url{https://www.aclweb.org/portal/content/acl-code-ethics}} We encourage all authors to include an explicit ethics statement on the broader impact of the work, or other ethical considerations after the conclusion but before the references. The ethics statement will not count toward the page limit (8 pages for long, 4 pages for short papers).

\section*{Limitations and Future Work}
The main limitation of our work is the limited sample size afforded by use of a physical robot. However, we believe the use of an embodied agent is essential for our data collection, as embodied agents provide experiences closer to that of natural human interaction than virtual interactions. Future work can use our system to replicate the data collection through a physically embodied robot.

Another limitation of our experimental results is that we only ablated contribution of modalities, but did not further interpret behavioral cues that might influence model performance. As such, these results are less interpretable due to lack of additional fine-grained annotations. Future work can obtain fine-grained annotations of the video data for empathy-relevant behavioral cues such as arousal, valence, self disclosure, etc. 

Our dataset is a valuable resource for furthering research in empathy modeling for AI systems. Novel future directions to explore could include personalized modeling of empathy patterns, using the longitudinal data as well as understanding cognitive insights behind when empathy arises in personal story sharing.

% \section*{Acknowledgements}
% This document has been adapted by Jordan Boyd-Graber, Naoaki Okazaki, Anna Rogers from the style files used for earlier ACL, EMNLP and NAACL proceedings, including those for
% EACL 2023 by Isabelle Augenstein and Andreas Vlachos,
% EMNLP 2022 by Yue Zhang, Ryan Cotterell and Lea Frermann,
% ACL 2020 by Steven Bethard, Ryan Cotterell and Rui Yan,
% ACL 2019 by Douwe Kiela and Ivan Vuli\'{c},
% NAACL 2019 by Stephanie Lukin and Alla Roskovskaya, 
% ACL 2018 by Shay Cohen, Kevin Gimpel, and Wei Lu, 
% NAACL 2018 by Margaret Mitchell and Stephanie Lukin,
% Bib\TeX{} suggestions for (NA)ACL 2017/2018 from Jason Eisner,
% ACL 2017 by Dan Gildea and Min-Yen Kan, NAACL 2017 by Margaret Mitchell, 
% ACL 2012 by Maggie Li and Michael White, 
% ACL 2010 by Jing-Shin Chang and Philipp Koehn, 
% ACL 2008 by Johanna D. Moore, Simone Teufel, James Allan, and Sadaoki Furui, 
% ACL 2005 by Hwee Tou Ng and Kemal Oflazer, 
% ACL 2002 by Eugene Charniak and Dekang Lin, 
% and earlier ACL and EACL formats written by several people, including
% John Chen, Henry S. Thompson and Donald Walker.
% Additional elements were taken from the formatting instructions of the \emph{International Joint Conference on Artificial Intelligence} and the \emph{Conference on Computer Vision and Pattern Recognition}.

\bibliography{custom}
\bibliographystyle{acl_natbib}

% \clearpage
% \newpage

\begin{figure*}[ht]
\centering
         \includegraphics[width=\textwidth]{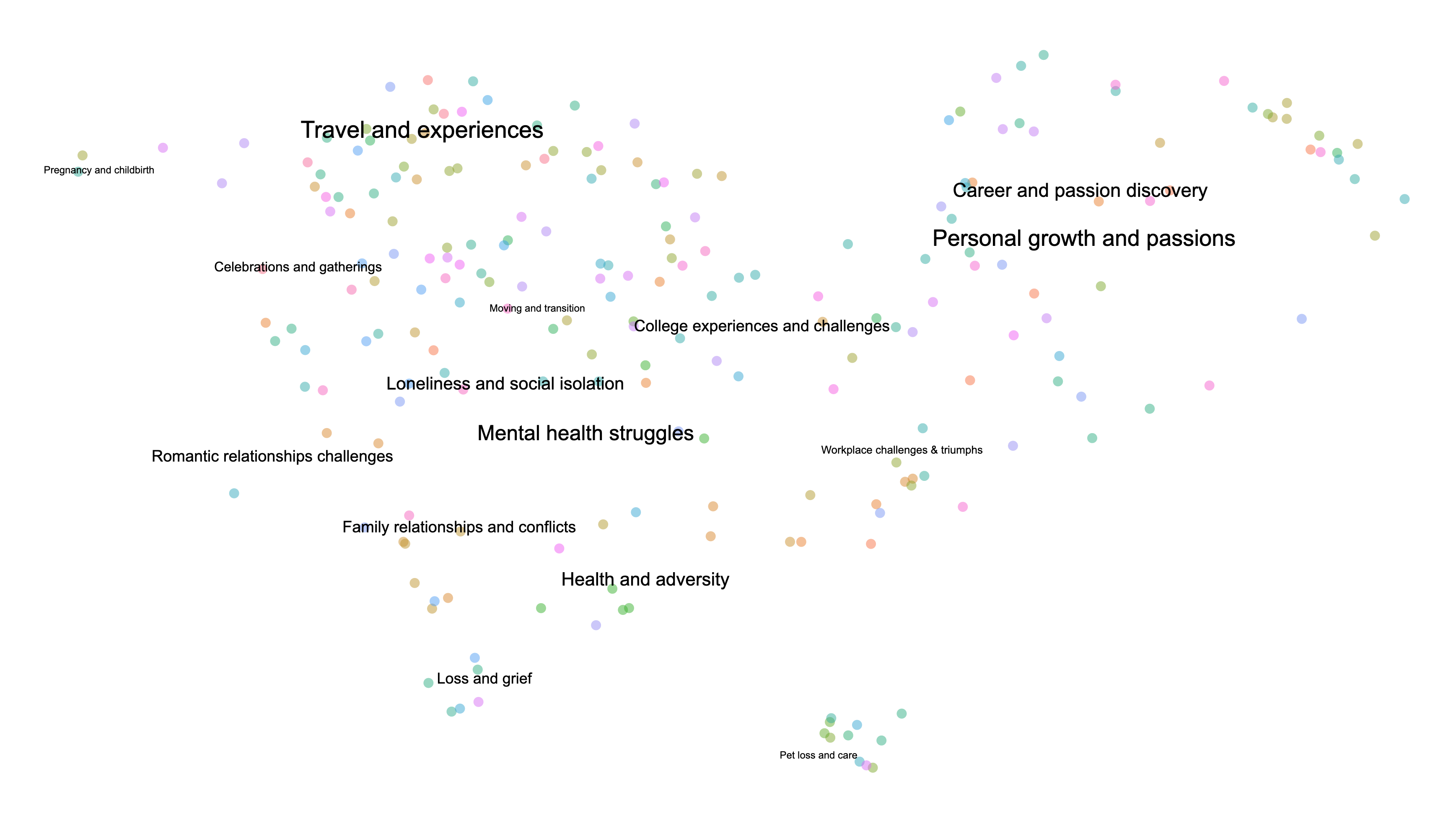}
         \caption{\textbf{Story Topics}: We visualize the embeddings  (obtained with UMAP of ada-002 embeddings) of Story topics. Our deployment across the United States gives us a diverse set of meaningful personal stories. 
         % This enables xyz research.
         }
         % \caption{Story topics participants shared (obtained with UMAP of ada-002 embeddings)}
         \label{fig:topics}
\end{figure*}

\appendix

\section{Story Topics} \label{appendix:topics}
The topics in Figure \ref{fig:topics} were obtained as follows: ada-v002 embeddings of stories were calculated via the OpenAI API, and a UMAP model was fit on the data to reduce the 1536 dimension vectors to $x$ and $y$ coordinates using cosine similarity as the distance measure and clusters were obtained with K-means.

\section{Implementation Details} \label{implementation}

We train our models on 4 NVIDIA RTX A6000 with a batch size of 64 for 10 epochs. We use the AdamW \cite{loshchilov2019decoupled} optimizer with an initial learning rate of 1e-4 with a scheduler $\text{StepLR}$ that decays the learning rate by 0.1 ($\gamma$) every 5 epochs (step\_size). For the loss function, we use the Mean Squared Error (MSE). For the dataloader, we first conduct oversampling based on the empathy ratings due to its imbalance distribution as shown in \Cref{fig:basicstatsc}. Next, we separate participants into train/valid/test sets in the ratio of 0.7/0.2/0.1 to ensure the model does not see the participants who were in the train sets. All models except for GPT-4 and EmpatheticStoriesBART were re-implemented to output multimodal representations that can be used to calculate similarities of story embeddings. We follow the default model parameters from the original implementations. 

\section{Prompting} \label{prompts}
We include prompts for GPT-4 benchmarking below:

\noindent\textbf{Story Sharing:}
\begin{itemize}
    \item \textbf{System prompt:} 
        \textit{You are a psychologist with expertise in analyzing empathy.
        You can predict how much people might empathize with each other, based on their past experiences.
        }

    \item \textbf{User prompt:}
        \textit{You will receive two stories, one from person A and the other from person B.
        Please predict, on a scale from 0 to 1, how much person A would empathize with B's story.
        Return just the number, no other text.
        }
\end{itemize}

\noindent\textbf{Reflection:}
\begin{itemize}
    \item \textbf{System prompt:} 
        \textit{You are a psychologist with expertise in analyzing empathy.
        You can predict how much people might empathize with each other, based on their past experiences.
        }

    \item \textbf{User prompt:}
        \textit{You will receive a story and conversation between person A and person B about person A's reflections about the story.
        Based on this, please predict, on a scale from 0 to 1, how much person A would empathize with the story.
        Return just the number, no other text.
        }
\end{itemize}

% \section{Example Appendix}
% \label{sec:appendix}

% This is a section in the appendix.

\end{document}